\documentclass[conference]{IEEEtran}
\IEEEoverridecommandlockouts
\usepackage{cite}
\usepackage{amsmath,amssymb,amsfonts}
\usepackage{algorithmic}
\usepackage{algorithm}
\usepackage{graphicx}
\usepackage{textcomp}
\usepackage{xcolor}
\usepackage{booktabs}
\usepackage{array}
\usepackage{multirow}
\usepackage{hyperref}
\usepackage{url}
\usepackage{balance}
\usepackage{enumitem}
\usepackage{pifont}
\usepackage{flafter}   % floats always appear after first reference
\usepackage{placeins}  % provides \FloatBarrier
\usepackage{float}     % enables [H] exact-placement specifier

\def\BibTeX{{\rm B\kern-.05em{\sc i\kern-.025em b}\kern-.08em
    T\kern-.1667em\lower.7ex\hbox{E}\kern-.125emX}}

\begin{document}

\title{CLEAR-MoE: Shared-Basis Expert Extraction from Frozen Vision Transformers via Calibration-Driven Layer Selection}

\author{\IEEEauthorblockN{Md Irtiza Hossain}
\IEEEauthorblockA{\textit{Brac University}\\
mohammad.irtiza.hossain@gmail.com}
\and
\IEEEauthorblockN{Humaira Ayesha}
\IEEEauthorblockA{\textit{Brac University}\\
humaira.ayeshaaa@gmail.com}
\and
\IEEEauthorblockN{Junaid Ahmed Sifat}
\IEEEauthorblockA{\textit{Brac University}\\
junaid.ahmed.sifat@g.bracu.ac.bd}}

\maketitle

%───────────────────────────────────────────────────────────────────────────────────
\begin{abstract}
We present \textbf{CLEAR-MoE}, a four-phase post-training pipeline that converts a frozen pretrained vision transformer (ViT) into a sparse mixture-of-experts (MoE) model without updating backbone weights. The pipeline (i)~scores FFN layers by sparsity, clusterability, and output sensitivity; (ii)~decomposes selected layers into a shared low-rank SVD basis plus per-cluster residual experts ($k$-means); (iii)~fits lightweight routers supervised by cluster labels; and (iv)~dispatches tokens via pluggable CUDA backends. On Imagenette with DeiT-Small, CLEAR-MoE retains 99.9\% of dense accuracy (86.70\,$\pm$\,0.02\% vs.\ 86.73\%). Our ablations isolate a consistent empirical finding: the shared SVD basis is the dominant accuracy-preserving component. Random routing, learned routing, and three router architectures yield numerically similar results spanning at most 0.06\,pp (86.62--86.68\%), and accuracy remains stable across SVD rank, expert count $E \in \{2,\ldots,8\}$, calibration size $N\in\{50,\ldots,500\}$, and random seed. This finding generalizes across five ViT backbones (DeiT-T/S/B, ViT-S/B, 5.7M--86.6M parameters), with $|\Delta|{\leq}0.10$\,pp across all configurations. On a GTX\,960, routing and scatter-gather overhead make CLEAR-MoE FFN 1.3--1.7$\times$ slower than dense. A dispatch microbenchmark indicates that routing (AI\,=\,1.9\,FLOPs/B) is an order of magnitude more memory-bound than expert GEMMs (AI\,=\,22.7), identifying fused dispatch kernels as a plausible optimization target.
\end{abstract}


%â”€â”€â”€â”€â”€â”€â”€â”€â”€â”€â”€â”€â”€â”€â”€â”€â”€â”€â”€â”€â”€â”€â”€â”€â”€â”€â”€â”€â”€â”€â”€â”€â”€â”€â”€â”€â”€â”€â”€â”€â”€â”€â”€â”€â”€â”€â”€â”€â”€â”€â”€â”€â”€â”€â”€â”€â”€â”€â”€â”€â”€â”€â”€â”€â”€â”€â”€â”€â”€â”€â”€â”€â”€â”€â”€â”€â”€
\section{Introduction}
\label{sec:intro}

Vision transformers (ViTs) \cite{dosovitskiy2021vit} devote roughly two-thirds of inference FLOPs to feed-forward network (FFN) sub-layers that execute the same MLP computation for every token, regardless of semantic content \cite{riquelme2021vmoe}. Sparse mixture-of-experts (MoE) architectures \cite{shazeer2017outrageously} address this by routing each token to a subset of specialized sub-networks, reducing active compute proportional to $k/E$. However, existing vision MoE models \cite{riquelme2021vmoe,chen2023adamvmoe} require training from scratch, which is a resource barrier for practitioners who already hold a pretrained checkpoint.

\emph{Post-training expert extraction} converts a dense FFN into experts without retraining \cite{zhang2022moefication,d2dmoe2024,berisha2025cvpr}. Despite recent progress, three gaps remain: (1)~most methods convert all FFN layers, ignoring sensitivity variation across depth; (2)~disjoint expert assignment discards shared low-level structure, destabilizing downstream transfer; (3)~speedup claims are reported in MACs rather than wall-clock latency on real hardware.

CLEAR-MoE addresses all three with the following contributions:

\begin{itemize}[leftmargin=*,noitemsep]
\item \textbf{Calibration-driven layer scoring}: composite score $\mathcal{S}(l) = 0.4\,\text{sparsity} + 0.4\,\text{clusterability} - 0.2\,\text{sensitivity}$ selects FFN layers that cluster well and tolerate perturbation, automatically excluding high-sensitivity layers (e.g., block~0, sensitivity~$=$~0.946).

\item \textbf{Shared-basis decomposition}: fc2 is decomposed into a shared truncated-SVD basis plus per-cluster residuals; fc1 is shared identically across all experts. This preserves common visual structure while allowing per-cluster specialization.

\item \textbf{Hardware-transparent latency study}: all timing on a single GTX\,960 (4\,GB, 112\,GB/s) with \texttt{cuda.synchronize()}-fenced p50 reporting (100 forward passes), indicating that routing overhead (not expert arithmetic) is a likely bottleneck.

\item \textbf{Comprehensive ablation}: decomposition (D0--D8), layer selection (L0--L10), dispatch backends, expert count, SVD rank, router architecture, calibration size, and random seed, establishing which design choices actually matter.
\end{itemize}

\noindent\textbf{Scope.} All measured results use a single GTX\,960. No claims are made about data-center GPUs, distributed setups, or larger datasets. Multi-device throughput projections are modelled analytically using PCIe Gen3$\times$16 bandwidth; no NCCL runs were performed.

%â”€â”€â”€â”€â”€â”€â”€â”€â”€â”€â”€â”€â”€â”€â”€â”€â”€â”€â”€â”€â”€â”€â”€â”€â”€â”€â”€â”€â”€â”€â”€â”€â”€â”€â”€â”€â”€â”€â”€â”€â”€â”€â”€â”€â”€â”€â”€â”€â”€â”€â”€â”€â”€â”€â”€â”€â”€â”€â”€â”€â”€â”€â”€â”€â”€â”€â”€â”€â”€â”€â”€â”€â”€â”€â”€â”€â”€
\section{Related Work}
\label{sec:related}

\textbf{Post-training expert extraction.} MoEfication \cite{zhang2022moefication} first showed that clustering neuron activation patterns partitions a pretrained FFN into experts without retraining. D2DMoE \cite{d2dmoe2024} extended this to dynamic-$k$ routing, achieving 30\% latency reduction on an A100 for ViT-B/ImageNet-1K. Berisha et al.\ \cite{berisha2025cvpr} used variance-based neuron grouping to recover 98\% of dense performance with 36.3\% MAC reduction for DeiT-B. CLEAR-MoE differs by (a)~selecting layers via a composite sensitivity-aware score, (b)~preserving shared structure via SVD decomposition, and (c)~providing a full wall-clock dispatch study on consumer hardware. Table~\ref{tab:priorwork} positions these methods side-by-side. Sparse Upcycling \cite{komatsuzaki2023sparse} converts dense language-model checkpoints to MoE by cloning FFN weights and training a load-balancing router via continued pre-training; unlike CLEAR-MoE, backbone weights are updated, requiring gradient access and sufficient unlabelled data.

\begin{table}[!t]
\caption{Post-training MoE extraction: method comparison. $\dagger$ = MAC reduction (not wall-clock). $\ddagger$ = minimal fine-tuning reported by original paper. CLEAR-MoE: zero backbone weight updates.}
\label{tab:priorwork}
\centering
\footnotesize
\setlength{\tabcolsep}{2.0pt}
\begin{tabular}{lp{1.4cm}lrrp{1.5cm}}
\toprule
\textbf{Method} & \textbf{Backbone} & \textbf{HW} & \textbf{Acc.} & \textbf{Latency} & \textbf{Structure} \\
\midrule
MoEfication \cite{zhang2022moefication}  & ViT  & N/A       & $\sim$99\%     & N/A                 & Disjoint $k$-means \\
D2DMoE \cite{d2dmoe2024}$^{\ddagger}$    & ViT-B/16 & A100  & $\sim$99\%     & $-$30\%             & Dynamic-$k$ disj. \\
Berisha \cite{berisha2025cvpr}$^{\ddagger}$ & DeiT-B & N/A  & 98\%           & $-$36.3\%$\dagger$  & Var.-based disj. \\
\textbf{CLEAR-MoE}                      & \textbf{DeiT/ViT (5)} & \textbf{GTX\,960} & \textbf{99.9\%} & \textbf{$+$1.3--1.7$\times$} & \textbf{Shared + res.} \\
\bottomrule
\multicolumn{6}{p{8.2cm}}{\footnotesize D2DMoE latency on A100 (2\,TB/s); CLEAR-MoE on GTX\,960 (112\,GB/s). Bandwidth 18$\times$ lower.}
\end{tabular}
\end{table}

\textbf{Vision MoE training.} V-MoE \cite{riquelme2021vmoe} demonstrated that sparse patch routing matches dense quality at $\sim$50\% active compute for 15B-parameter ViTs. AdaMV-MoE \cite{chen2023adamvmoe} showed late transformer layers benefit most from expertization, a heuristic our composite score can override when sensitivity signals contradict it. DynamicViT \cite{rao2021dynamicvit} and A-ViT \cite{yin2022avit} achieve token-level sparsification via halting gates, complementary to our FFN-level approach.

\textbf{MoE runtime systems.} TUTEL \cite{hwang2023tutel} achieves 3.11$\times$ speedup via 2D all-to-all on 128 GPUs. Brainstorm \cite{brainstorm2023} shows profile-guided dispatch yields 5$\times$ speedup over DeepSpeed for SwinV2-MoE. Lancet \cite{lancet2024} reduces non-overlapping communication by 77\% via whole-graph overlap. These systems target high-bandwidth multi-GPU clusters; CLEAR-MoE studies the single-consumer-GPU regime where bandwidth is 18$\times$ lower.

%â”€â”€â”€â”€â”€â”€â”€â”€â”€â”€â”€â”€â”€â”€â”€â”€â”€â”€â”€â”€â”€â”€â”€â”€â”€â”€â”€â”€â”€â”€â”€â”€â”€â”€â”€â”€â”€â”€â”€â”€â”€â”€â”€â”€â”€â”€â”€â”€â”€â”€â”€â”€â”€â”€â”€â”€â”€â”€â”€â”€â”€â”€â”€â”€â”€â”€â”€â”€â”€â”€â”€â”€â”€â”€â”€â”€â”€
\section{Method}
\label{sec:method}

\begin{figure}[!t]
\centering
\includegraphics[width=\columnwidth]{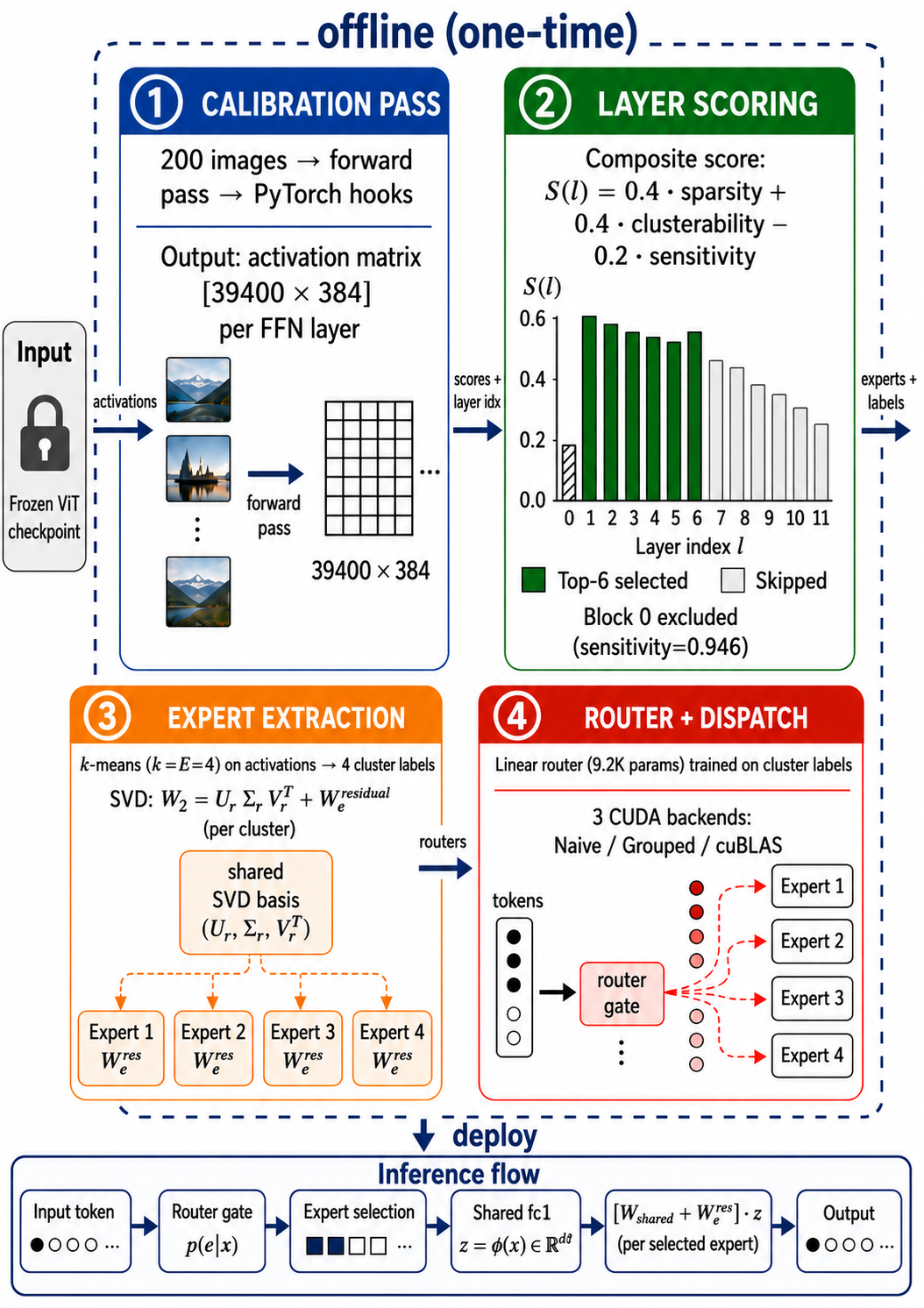}
\caption{CLEAR-MoE pipeline. A frozen pretrained ViT yields calibration activations; scored FFN layers are decomposed into a shared SVD basis plus per-cluster residual experts; lightweight routers are fitted on $k$-means labels; inference uses pluggable CUDA dispatch backends.}
\label{fig:pipeline}
\end{figure}

CLEAR-MoE converts a frozen pretrained ViT into a selectively expertised model through four sequential phases (Fig.~\ref{fig:pipeline}). No backbone weights are updated.

\subsection{Phase 1: Calibration Pass}

Forward $N_\text{cal}$ representative images through the frozen model, recording pre-FFN activation tensors at each of the $L$ selected layers via PyTorch hooks. For DeiT-Small ($d{=}384$, $N_\text{cal}{=}200$, 197 tokens/image), this produces a $[39{,}400,\,384]$ activation matrix per layer.

\subsection{Phase 2: Layer Scoring and Selection}

Each FFN layer $l$ receives a composite score:
\begin{equation}
\mathcal{S}(l) = 0.4\,\text{sparsity}(l) + 0.4\,\text{clusterability}(l) - 0.2\,\text{sensitivity}(l)
\label{eq:score}
\end{equation}
where \emph{sparsity} is the fraction of activation magnitudes below 0.01; \emph{clusterability} is the Silhouette score of $k$-means with $k{=}E$ on the calibration activations; and \emph{sensitivity} is the normalized logit change when the layer's FFN output is zeroed. Sensitivity is \emph{subtracted} because high sensitivity implies high degradation risk if expertized. The top-$k{=}L/2$ layers by $\mathcal{S}$ are selected; the remaining layers retain their original dense FFN. Table~\ref{tab:layer_scores_main} reports per-layer scores for DeiT-Small (200-image calibration); block~0 ranks last due to sensitivity~$=$~0.946.

\begin{table}[!t]
\caption{Per-layer composite scores, DeiT-Small (200-image calibration). Composite $= 0.4{\cdot}\text{sp} + 0.4{\cdot}\text{cl} - 0.2{\cdot}\text{se}$. Bold = selected (top-$k{=}6$).}
\label{tab:layer_scores_main}
\centering
\small
\resizebox{\columnwidth}{!}{%
\begin{tabular}{lrrrr}
\toprule
\textbf{Block} & \textbf{Sparsity} & \textbf{Clusterab.} & \textbf{Sensitivity} & \textbf{Composite} \\
\midrule
\textbf{blocks.1} & 0.180 & 0.526 & 0.171 & \textbf{0.248} \\
\textbf{blocks.4} & 0.152 & 0.518 & 0.109 & \textbf{0.246} \\
\textbf{blocks.5} & 0.152 & 0.520 & 0.113 & \textbf{0.246} \\
\textbf{blocks.3} & 0.149 & 0.519 & 0.113 & \textbf{0.245} \\
\textbf{blocks.6} & 0.147 & 0.522 & 0.123 & \textbf{0.243} \\
\textbf{blocks.2} & 0.141 & 0.518 & 0.147 & \textbf{0.234} \\
blocks.7  & 0.147 & 0.519 & 0.162 & 0.234 \\
blocks.11 & 0.134 & 0.516 & 0.149 & 0.230 \\
blocks.9  & 0.135 & 0.518 & 0.167 & 0.228 \\
blocks.8  & 0.142 & 0.512 & 0.175 & 0.227 \\
blocks.10 & 0.171 & 0.514 & 0.252 & 0.224 \\
blocks.0  & 0.210 & 0.516 & 0.946 & 0.101 \\
\bottomrule
\end{tabular}
}
\end{table}

\subsection{Phase 3: Expert Extraction}

For each selected layer, fc2 ($W_2 \in \mathbb{R}^{d \times d_\text{ffn}}$) is decomposed via truncated SVD; fc1 is shared identically across all experts:
\begin{equation}
y = \underbrace{W_\text{shared}}_{\text{SVD-}r\text{ approx.}} \!\cdot z + \underbrace{W_{e^*}^\text{res}}_{\text{cluster residual}} \!\cdot z, \;\; z = \text{GELU}(\text{fc1}(x))
\label{eq:decomp}
\end{equation}
$W_\text{shared} = U_r\Sigma_r V_r^\top$ retains the top-$r$ singular values (default $r$ = 50\% of matrix rank, $r{=}d/2{=}192$ for DeiT-Small where $W_2{\in}\mathbb{R}^{384\times1536}$ has maximum rank 384; materialised as a dense matrix before inference, with no factorised compute reduction). $k$-means with $k{=}E$ partitions calibration tokens into $E$ clusters; the per-cluster residual is $W_e^\text{res} = (W_2 - W_\text{shared})\cdot s_e$, where $s_e$ scales by the ratio of cluster-mean to global-mean activation norm. Because $s_e$ is a scalar, all residual experts share the same directional component $(W_2 - W_\text{shared})$, differing only in magnitude; a misrouted token therefore receives a different scaling but the same directional correction. During inference, $z$ is computed once and reused for both the shared and residual paths.

\subsection{Phase 4: Router Fitting}

A router $g_\theta$ predicts the cluster assignment of each token. The $k$-means labels from Phase~3 provide supervised targets: router training minimizes cross-entropy between $g_\theta(x)$ and cluster label $c$ over the calibration tokens, using AdamW ($\text{lr}{=}10^{-3}$, 5 epochs, cosine decay). Three router architectures are evaluated: \textbf{Linear} ($d\times E$ parameters, 9.2K total), \textbf{MLP} (hidden layer $d/6$, 149K), and \textbf{Adaptive} (confidence-threshold top-1/top-2 escalation, same parameter count as Linear).

\subsection{Dispatch Backends}

Three CUDA backends are implemented: (1)~\textbf{Naive}: boolean-mask loop over experts (reference). (2)~\textbf{Grouped}: tokens are sorted by expert index; one GEMM per contiguous sub-batch. No padding; throughput stable under imbalance. (3)~\textbf{cuBLAS}: expert sub-batches are padded to the largest and stacked into a 3D tensor dispatched to a batched GEMM call. Fastest at balanced load; degrades at high imbalance from padding waste.

%â”€â”€â”€â”€â”€â”€â”€â”€â”€â”€â”€â”€â”€â”€â”€â”€â”€â”€â”€â”€â”€â”€â”€â”€â”€â”€â”€â”€â”€â”€â”€â”€â”€â”€â”€â”€â”€â”€â”€â”€â”€â”€â”€â”€â”€â”€â”€â”€â”€â”€â”€â”€â”€â”€â”€â”€â”€â”€â”€â”€â”€â”€â”€â”€â”€â”€â”€â”€â”€â”€â”€â”€â”€â”€â”€â”€â”€
\section{Experiments}
\label{sec:results}

\textbf{Setup.} DeiT-Small (22M parameters, $d{=}384$, $d_\text{ffn}{=}1536$, 12 blocks) evaluated on Imagenette (3,925 validation images, 10-class ImageNet subset) with 200-image calibration set, $E{=}4$ experts, AdamW router training (5 epochs, lr\,$10^{-3}$, cosine decay). All timing: p50 of 100 forward passes, batch size 8, GTX\,960 (4\,GB, 112\,GB/s), \texttt{cuda.synchronize()}-fenced.

\subsection{Decomposition Ablation (D0--D8)}

Table~\ref{tab:cls} ablates what each architectural choice contributes to accuracy and latency. Configurations D0--D8 span the design space from dense baseline to shared-basis CLEAR-MoE FFN, isolating the effect of the shared basis, the routing mechanism, and the expert assignment strategy.

\begin{table}[!t]
\caption{Decomposition ablation (DeiT-Small, Imagenette, $E{=}4$, last-$k$ layers, batch=8, GTX\,960). $\Delta$Top-1 relative to D0. D4/D5 differ only in random seed. D6 and D8 are the same architecture: D6 is the cold-kernel first dispatch; D8 is measured after 3 warm-up passes (steady-state). The 21\,ms gap reflects CUDA JIT overhead on first run.}
\label{tab:cls}
\centering
\footnotesize
\setlength{\tabcolsep}{2.0pt}
\begin{tabular}{clllrrr}
\toprule
\textbf{ID} & \textbf{Sh.\ fc1} & \textbf{Sh.\ fc2} & \textbf{Router} & \textbf{Top-1} & $\boldsymbol{\Delta}$\textbf{pp} & \textbf{ms (p50)} \\
\midrule
D0 & -- & -- & Dense & 86.73\% & $+$0.00 & 59.6 \\
\midrule
D1 & \checkmark & SVD rank-$r$ & None & 86.55\% & $-$0.18 & 59.7 \\
D2 & \ding{55} & Disjoint & Random & 66.42\% & $-$20.31 & 49.4 \\
D3 & \checkmark & Full res. & None & 86.73\% & $+$0.00 & 59.9 \\
D4/D5 & \checkmark & $k$-means res. & Random & 86.62--65\% & $-$0.08--11 & 92.2 \\
D6/D8 & \checkmark & $k$-means res. & Linear & 86.65\% & $-$0.08 & 78.8--100.3 \\
D7 & \ding{55} & Disjoint & $k$-means & 65.22\% & $-$21.51 & 48.0 \\
\bottomrule
\multicolumn{7}{p{8.2cm}}{\footnotesize \ding{55}\,=\,disjoint (no shared fc1); \checkmark\,=\,shared across all tokens.}
\end{tabular}
\end{table}

\textbf{The shared SVD basis is the dominant accuracy-preserving component.} D3 (shared basis + full global residual, no routing) achieves exact reconstruction: algebraically, $W_\text{shared}{\cdot}z + (W_2{-}W_\text{shared}){\cdot}z = W_2{\cdot}z$, matching D0 at 86.73\%. D2 and D7 (disjoint experts, no shared fc1) collapse to 65--66\% regardless of routing strategy: disjoint expert weights estimated from 200 calibration images fail to generalize. In contrast, D4/D5 (shared basis, \emph{random} routing) retain 86.62--65\%, only 0.08--0.11\,pp below D0; D6/D8 (shared basis, learned routing at 95\% accuracy) achieve an identical $-$0.08\,pp. Routing quality has limited measured impact once the shared basis is present.

\textbf{No latency reduction on GTX\,960.} CLEAR-MoE FFN (D4--D8) executes shared fc1+fc2 on \emph{all} tokens plus residual paths that collectively process all $T$ tokens ($T/E$ per expert $\times$ $E$ experts), adding approximately one full fc2-equivalent computation; total arithmetic is $\approx\!1.5\times$ dense FFN. On bandwidth-limited memory (112\,GB/s), this incurs 1.3--1.7$\times$ latency overhead (78--100\,ms vs.\ 59.6\,ms). Only disjoint experts (D2/D7) are faster (48--49\,ms) by eliminating the shared path entirely, at the cost of 21\,pp accuracy.

\subsection{Layer-Selection Ablation (L0--L10)}
\label{sec:ablation}

Table~\ref{tab:ablation} compares all 11 layer-selection strategies on the same MoE configuration ($E{=}4$, $k{=}6$ of 12 FFN layers, composite scoring).

\begin{table}[!t]
\caption{Complete layer-selection ablation (L0--L10). All: DeiT-Small, $E{=}4$, $k{=}6$ layers, linear router, 200-image calibration. L1 avg of 3 seeds. $\Delta$ relative to L0.}
\label{tab:ablation}
\centering
\small
\setlength{\tabcolsep}{3pt}
\resizebox{\columnwidth}{!}{%
\begin{tabular}{llllrr}
\toprule
\textbf{ID} & \textbf{Strategy} & \textbf{Layers} & \textbf{Top-1} & $\boldsymbol{\Delta}$\textbf{pp} & \textbf{p50 ms} \\
\midrule
L0  & Dense (no MoE)    & --           & 86.73\%          & $+$0.00 & 57.0 \\
L1  & Random ($n{=}3$)  & varies       & 86.72$\pm$0.01\% & $-$0.01 & 78.5$\pm$0.8 \\
L2  & First $k$         & 0--5         & 86.60\%          & $-$0.13 & 79.3 \\
L3  & Last $k$          & 6--11        & 86.68\%          & $-$0.05 & 80.3 \\
L4  & Alternating (odd) & 1,3,5,7,9,11 & 86.83\%         & $+$0.10 & 87.2 \\
L5  & Sparsity only     & 0,1,3,4,5,10 & 86.73\%         & $+$0.00 & 99.1 \\
L6  & Clusterability    & 1,2,3,5,6,7  & 86.73\%         & $+$0.00 & 88.8 \\
L7  & High-sensitivity  & 0,1,7,8,9,10 & 86.65\%         & $-$0.08 & 102.5 \\
L8  & Sp.+Cl.           & 0,1,4,5,6,10 & 86.62\%         & $-$0.10 & 88.8 \\
L9  & Cl.$-$Se.         & 2,3,4,5,6,11 & 86.68\%         & $-$0.05 & 94.0 \\
\textbf{L10} & \textbf{Composite (ours)} & \textbf{1--6} & \textbf{86.68\%} & $\boldsymbol{-}$\textbf{0.05} & \textbf{82.4} \\
\bottomrule
\end{tabular}
}
\end{table}

All 11 policies span only 0.21\,pp (86.62--86.83\%), confirming that \emph{accuracy is policy-insensitive}. The shared basis preserves quality regardless of which 6 blocks are expertized. The composite score's practical value is principled block~0 exclusion (sensitivity\,=\,0.946): L2 (first-$k$) includes block~0 and incurs the worst $-$0.13\,pp drop. L10 avoids it, achieving 82.4\,ms, comparable to L3 (last-$k$, 80.3\,ms) while additionally excluding high-sensitivity blocks. L7 (high-sensitivity, deliberately selecting the most sensitive blocks as a stress-test baseline) achieves only $-$0.08\,pp accuracy loss but is the slowest policy at 102.5\,ms, demonstrating that layer choice affects latency more than accuracy.

\subsection{Dispatch Strategy Benchmark}

We benchmark three CUDA backends ($B{=}8$, $T{=}1568$, $E{=}4$, GTX\,960) under four imbalance levels (Table~\ref{tab:dispatch_main}). At balanced load, cuBLAS peaks at 728\,K\,tok/s (2.79$\times$ CPU serial). At 80\% imbalance, padding waste collapses cuBLAS to 558\,K\,tok/s ($-$23\%). Grouped dispatch remains stable at 665--747\,K\,tok/s across all imbalance levels. \textbf{Recommendation}: cuBLAS for predictably balanced load; Grouped otherwise.

\begin{table}[!t]
\caption{Dispatch micro-benchmark: 0\% vs.\ 80\% imbalance ($B{=}8$, $T{=}1568$, $E{=}4$, GTX\,960).}
\label{tab:dispatch_main}
\centering
\small
\setlength{\tabcolsep}{6pt}
\begin{tabular}{lrrrr}
\toprule
& \multicolumn{2}{c}{\textbf{0\% imb.}} & \multicolumn{2}{c}{\textbf{80\% imb.}} \\
\cmidrule(lr){2-3}\cmidrule(lr){4-5}
\textbf{Backend} & ms & K\,tok/s & ms & K\,tok/s \\
\midrule
CPU Serial & 6.01 & 261 & 5.96 & 263 \\
Naive      & 3.67 & 428 & 3.59 & 437 \\
Grouped    & 2.36 & \textbf{665} & 2.14 & \textbf{733} \\
cuBLAS     & \textbf{2.15} & \textbf{728} & 2.81 & 558 \\
\bottomrule
\end{tabular}
\end{table}

\subsection{Roofline Analysis}

\begin{figure}[!t]
\centering
\includegraphics[width=\columnwidth]{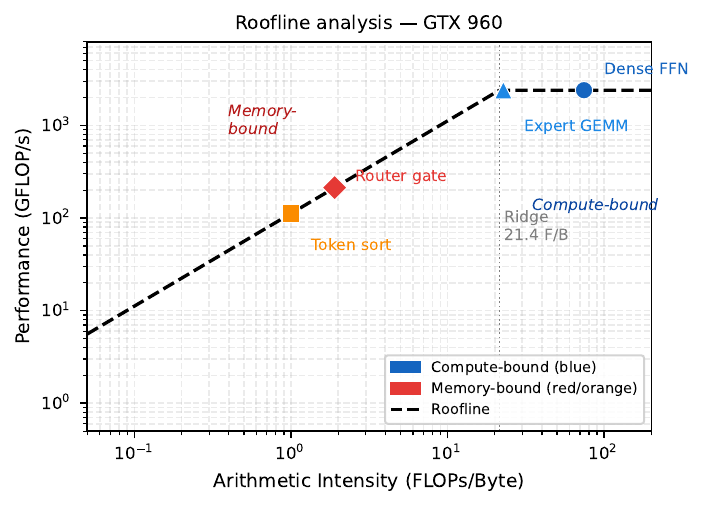}
\caption{Roofline for GTX\,960. Blue = compute-bound; orange/red = memory-bound. Router gate (AI\,=\,1.9) is 11$\times$ below the ridge point (21.4\,FLOPs/B); routing, not expert arithmetic, is the primary latency bottleneck.}
\label{fig:roofline}
\end{figure}

Fig.~\ref{fig:roofline} places each CLEAR-MoE operation on the GTX\,960 roofline (peak FP32: 2.4\,TFLOP/s, BW: 112\,GB/s, ridge: 21.4\,FLOPs/B). Dense FFN (AI\,=\,74.3) and expert GEMMs (AI\,=\,22.7) are compute-bound. The router gate (AI\,=\,1.9) and token sort (AI\,=\,1.0) are \emph{both} deep in the memory-bound regime: 11$\times$ and 21$\times$ below the ridge point. This indicates that routing and token sort are strongly memory-bound (11--21$\times$ below the ridge point), making fused dispatch kernels a plausible high-leverage optimization. Expert GEMMs are already near the compute ceiling and offer diminishing returns.

\subsection{Hyperparameter Sensitivity}

We ablate SVD rank, expert count, router architecture, calibration size, and random seed (Table~\ref{tab:sensitivity_main}); findings are:

\begin{table}[!t]
\caption{Hyperparameter sensitivity summary (DeiT-Small, Imagenette). All ranges $\leq$0.41\,pp.}
\label{tab:sensitivity_main}
\centering
\footnotesize
\setlength{\tabcolsep}{3pt}
\begin{tabular}{llrr}
\toprule
\textbf{Study} & \textbf{Range tested} & \textbf{Top-1 span} & $\boldsymbol{\Delta}$\textbf{pp} \\
\midrule
SVD rank $r$   & 16--256           & 86.62--86.75\%           & 0.13 \\
Expert count $E$ & 2--16           & 86.39--86.80\%$^*$       & 0.41 \\
Router arch.   & Lin/MLP/Adaptive  & \textbf{86.68\%} (all)   & 0.00 \\
Calibration $N$ & 50--500          & 86.55--86.70\%           & 0.15 \\
Random seed    & 42/123/456        & 86.70$\pm$0.02\%         & 0.05 \\
\bottomrule
\multicolumn{4}{p{8.2cm}}{\footnotesize $^*E{\in}\{2,4,8\}$ spans 0.15\,pp; $E{=}16$ drops 0.26\,pp (calibration instability: 200 images insufficient for 16 clusters).}
\end{tabular}
\end{table}

\textbf{SVD rank} ($r \in \{16, 32, 64, 96, 128, 192, 256\}$): accuracy range is 86.62--86.75\% (0.13\,pp). Reconstruction error decreases monotonically from 0.91 to 0.29, but this improvement does not translate to accuracy. Even $r{=}16$ captures the shared basis sufficiently.

\textbf{Expert count} ($E \in \{2, 4, 8, 16\}$): accuracy is 86.65--86.80\% for $E\in\{2,4,8\}$; $E{=}16$ drops 0.26\,pp as 200 calibration images provide insufficient statistics to estimate 16 fine-grained clusters. Latency grows $\sim$9\,ms per doubling of $E$ (73\,ms at $E{=}2$ to 98\,ms at $E{=}16$). No empty experts appear at any count.

\begin{figure}[!t]
\centering
\includegraphics[width=\columnwidth]{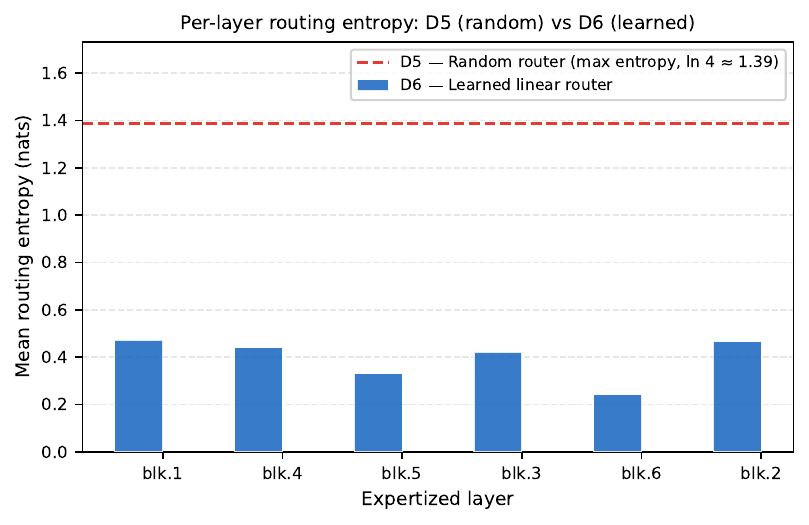}
\caption{Mean per-token routing entropy (nats) per expertized layer. Dashed red = D5 (random routing, max entropy $\ln 4{\approx}1.386$). Blue bars = D6 (learned linear router, mean ${\approx}0.40$\,nats). Despite $3.5\times$ entropy reduction, the D5-vs-D6 accuracy gap is only 0.06\,pp, confirming that routing confidence does not drive accuracy; the shared SVD basis does.}
\label{fig:expert_entropy}
\end{figure}

\textbf{Router architecture}: all three routers (Linear, MLP, Adaptive) achieve identical 86.68\% Top-1. MLP routing accuracy reaches 0.980 vs.\ 0.925 for Linear, a 5.5\,pp improvement in routing precision with zero downstream benefit. The gap between random routing (D4/D5, Table~\ref{tab:cls}) and all learned routers is at most 0.06\,pp (roughly two changed predictions out of 3,925). While this numerically exceeds the 3-seed standard deviation of $\pm$0.02\,pp (Table~\ref{tab:sensitivity_main}, row ``Random seed''), a paired significance test would be needed to confirm a reliable effect; we treat 0.06\,pp as practically negligible. Fig.~\ref{fig:expert_entropy} provides a mechanistic view: despite D6 (learned router) reducing mean per-token routing entropy $3.5\times$ versus D5 (random, $\ln 4{\approx}1.386$\,nats), the accuracy impact is negligible.

\begin{figure}[!t]
\centering
\includegraphics[width=\columnwidth]{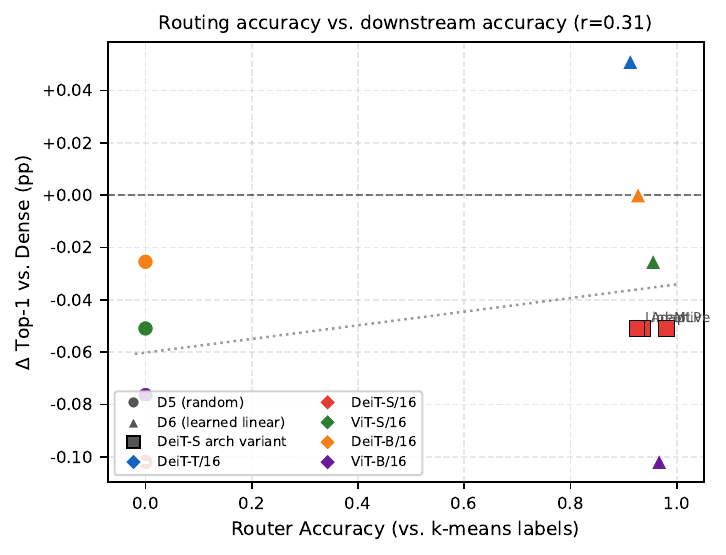}
\caption{Accuracy gap ($\Delta$pp relative to dense) vs.\ router training across 13 configurations. D5 (random, no trained router) at $x{=}0$; D6 and cross-backbone routers at measured routing accuracy. No strong monotonic relationship is visible, consistent with the hypothesis that shared-basis quality governs accuracy regardless of routing precision.}
\label{fig:router_scatter}
\end{figure}

Fig.~\ref{fig:router_scatter} plots accuracy gap against router training across 13 configurations (D5 at $x{=}0$, no trained router); no strong monotonic relationship is visible, consistent with the hypothesis that shared-basis quality governs accuracy regardless of routing quality.

\textbf{Calibration size} ($N \in \{50, 100, 200, 500\}$, multiple subsets for $N{<}500$): accuracy spans 86.55--86.70\% across all sizes; $N{=}50$ yields 86.65--86.70\% (within 0.15\,pp of $N{=}500$). Router accuracy climbs from 0.83 to 0.95 as $N$ grows, without accuracy benefit.

\textbf{Reproducibility}: 86.70\,$\pm$\,0.02\% Top-1 across seeds 42, 123, 456 (full pipeline, composite scoring, $E{=}4$, $N{=}200$).

\subsection{Cross-Backbone Generalization}
\label{sec:cross_backbone}

Table~\ref{tab:multibackbone} tests whether the shared-basis finding extends across architectures. We apply the full CLEAR-MoE pipeline to five backbones spanning 5.7--86.6\,M parameters and two pretraining lineages (DeiT, ViT); all other settings are identical to the DeiT-S ablation ($E{=}4$, $N{=}200$, seed\,42).

\begin{table}[!t]
\caption{Cross-backbone generalization (Imagenette, $E{=}4$, $N{=}200$, seed\,42). D5\,=\,random routing; D6\,=\,learned linear router. $|\Delta|{\leq}0.10$\,pp across all 10 configurations.}
\label{tab:multibackbone}
\centering
\small
\setlength{\tabcolsep}{3pt}
\resizebox{\columnwidth}{!}{%
\begin{tabular}{lrrrrrr}
\toprule
\textbf{Backbone} & \textbf{Params} & \textbf{Dense} & \textbf{D5\,$\Delta$\,pp} & \textbf{D6\,$\Delta$\,pp} & \textbf{D6 Rtr.\ Acc} & \textbf{D6 Skew} \\
\midrule
DeiT-T/16 & 5.7\,M  & 75.92\% & $-$0.05 & $+$0.05$^{\dagger}$ & 0.913 & 0.090 \\
DeiT-S/16 & 22.1\,M & 86.73\% & $-$0.10 & $-$0.05              & 0.925 & 0.099 \\
ViT-S/16  & 22.1\,M & 76.23\% & $-$0.05 & $-$0.03              & 0.956 & 0.139 \\
DeiT-B/16 & 86.6\,M & 91.77\% & $-$0.03 & $\pm$0.00            & 0.927 & 0.094 \\
ViT-B/16  & 86.6\,M & 85.38\% & $-$0.08 & $-$0.10$^{\ddagger}$ & 0.967 & 0.150 \\
\bottomrule
\multicolumn{7}{l}{\footnotesize $^{\dagger}$Single-seed result for this backbone; per-backbone seed variance not characterised.} \\
\multicolumn{7}{l}{\footnotesize $^{\ddagger}$Highest load skew (0.150); see Discussion.}
\end{tabular}
}
\end{table}

\begin{figure}[!t]
\centering
\includegraphics[width=\columnwidth]{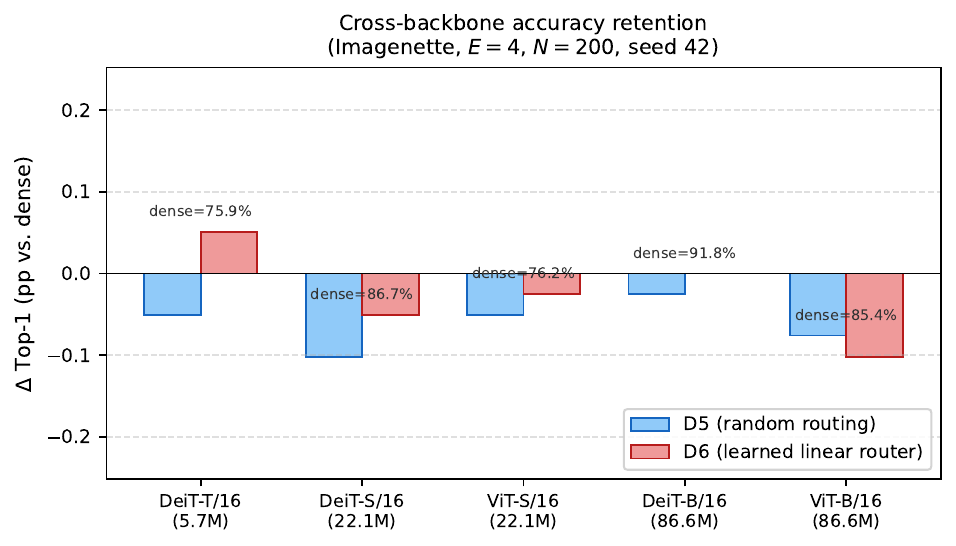}
\caption{Accuracy $\Delta$ vs.\ dense for five ViT backbones under D5 (random) and D6 (learned linear) routing (Imagenette, $E{=}4$, $N{=}200$, seed\,42). All deltas within $|\Delta|{\leq}0.10$\,pp. D6 provides marginal numerical advantage in 4 of 5 backbones; D5 outperforms D6 only on ViT-B (highest load skew, 0.150).}
\label{fig:backbone_comparison}
\end{figure}

\textbf{Routing differences are numerically small.} Learned routing (D6) provides a marginal numerical advantage over random routing (D5) in 4 of 5 backbones; however, the per-backbone improvements range from 0.02 to 0.10\,pp, too small to confirm without paired significance testing. The one exception is ViT-B/16, where D5 ($-$0.08\,pp) outperforms D6 ($-$0.10\,pp): The ViT-B result coincides with the highest observed load skew (0.150, vs.\ 0.094 for DeiT-B), which may contribute to the difference; no ablation was performed to confirm a causal link. Taken together, these results are consistent with the hypothesis that routing quality has limited impact once the shared SVD basis is present; a definitive claim would require paired prediction-level testing across additional seeds. Fig.~\ref{fig:backbone_comparison} visualises all 10 configurations.

%â”€â”€â”€â”€â”€â”€â”€â”€â”€â”€â”€â”€â”€â”€â”€â”€â”€â”€â”€â”€â”€â”€â”€â”€â”€â”€â”€â”€â”€â”€â”€â”€â”€â”€â”€â”€â”€â”€â”€â”€â”€â”€â”€â”€â”€â”€â”€â”€â”€â”€â”€â”€â”€â”€â”€â”€â”€â”€â”€â”€â”€â”€â”€â”€â”€â”€â”€â”€â”€â”€â”€â”€â”€â”€â”€â”€â”€
\section{Discussion}
\label{sec:discussion}

\textbf{Why routing differences are small.} All residual experts are scalar-conditioned variants of the same matrix $(W_2 - W_\text{shared})$, so mis-routing changes only the scaling factor applied to a token's correction, not its direction. The 0.06\,pp gap between random and learned routing is the direct consequence. Cross-backbone results (Table~\ref{tab:multibackbone}) are consistent across 4 of 5 architectures; a linear router ($<$10K parameters) appears adequate. Expert Choice routing \cite{zhou2022expertchoice}, where experts select their tokens rather than the reverse, achieves better load balance in language models; on DeiT-S, however, random and learned routing differ by at most 0.06\,pp, and all cross-backbone random-routing losses remain within 0.10\,pp of dense, supporting the hypothesis that routing strategy is secondary to decomposition quality.

\textbf{Why CLEAR-MoE is slower on the GTX\,960.} CLEAR-MoE FFN computes shared fc1+fc2 on \emph{all} tokens, then adds residual paths that each process $T/E$ tokens but collectively span all $T$ tokens across $E$ experts, adding approximately one full fc2-equivalent computation. Total arithmetic is $\approx\!1.5\times$ dense FFN. On a bandwidth-limited GPU (112\,GB/s vs.\ 2\,TB/s for an A100), loading shared weights and expert-residual weights across memory dominates. The dispatch micro-benchmark isolates this: routing and token sort (AI\,=\,1.0--1.9) are 11--21$\times$ below the compute ridge. Disjoint experts (D2/D7) are faster (48\,ms) by eliminating shared paths, but sacrifice 21\,pp accuracy. Analytical modelling suggests $\geq$800\,GB/s bandwidth may be needed for latency parity; A100 (2\,TB/s) or H100 (3.35\,TB/s) class hardware are the plausible candidates.

\textbf{Practical design guidelines.}
\emph{When to use CLEAR-MoE.} CLEAR-MoE may be well suited when accuracy preservation is non-negotiable and backbone fine-tuning is infeasible: the evaluated DeiT and ViT backbones can be converted with 200 calibration images and no GPU cluster. The method is not competitive with disjoint-expert approaches if latency reduction is the primary goal on consumer hardware.

\emph{Choosing $E$ and $r$.} Expert count $E{=}4$ provides a reasonable empirical trade-off: accuracy remains stable from $E{=}2$ to $E{=}8$, and $E{=}4$ balances router overhead with residual expressiveness. For $E{=}16$, our analytical model suggests $\geq$800 calibration images may be needed for stable cluster estimates; this was not directly validated. SVD rank $r$ can be set as low as 16 without consistent measured accuracy reduction on Imagenette; the default $r{=}d/2{\approx}192$ for DeiT-Small is conservative and may waste compute on deep layers.

\emph{Choosing the dispatch backend.} cuBLAS is optimal only for perfectly balanced expert load, which is common in classification but rare in detection/segmentation. For workloads with variable token-to-expert ratios, Grouped dispatch is approximately 31\% faster than cuBLAS at 80\% imbalance and remains comparatively stable across imbalance levels. Naive dispatch is a reference implementation only and is not recommended for latency-sensitive deployment.

\textbf{Limitations.}
\emph{Backbone and dataset scope.} DeiT-T/S/B and ViT-S/B generalization is confirmed (Table~\ref{tab:multibackbone}); ViT-L, ImageNet-1K, and hierarchical backbones (Swin, ConvNeXt) are not evaluated. Parameter count, memory footprint, and FLOPs versus a same-hardware baseline are not reported.

\emph{Hardware and compute.} CLEAR-MoE is 1.3--1.7$\times$ slower than dense on GTX\,960 (112\,GB/s). Latency parity on high-bandwidth hardware is modelled analytically only; multi-device projections are simulated (PCIe Gen3$\times$16), not measured NCCL runs.

%â”€â”€â”€â”€â”€â”€â”€â”€â”€â”€â”€â”€â”€â”€â”€â”€â”€â”€â”€â”€â”€â”€â”€â”€â”€â”€â”€â”€â”€â”€â”€â”€â”€â”€â”€â”€â”€â”€â”€â”€â”€â”€â”€â”€â”€â”€â”€â”€â”€â”€â”€â”€â”€â”€â”€â”€â”€â”€â”€â”€â”€â”€â”€â”€â”€â”€â”€â”€â”€â”€â”€â”€â”€â”€â”€â”€â”€
\section{Conclusion}
\label{sec:conclusion}

CLEAR-MoE demonstrates that post-training expert extraction preserves ${\geq}99.9$\% of dense ViT accuracy via a shared SVD basis plus per-cluster residuals, requiring only 200 calibration images and a single consumer GPU. This finding is consistent across five ViT backbones spanning 5.7--86.6\,M parameters and two pretraining lineages (Table~\ref{tab:multibackbone}), showing that the shared SVD basis is the dominant accuracy-preserving factor while routing quality, SVD rank, expert count ($E\in\{2,4,8\}$), and calibration size ($N\in\{50,\ldots,500\}$) are secondary. Random routing achieves 86.62\% while learned routing at 98\% accuracy achieves 86.68\% (a 0.06\,pp difference), simplifying the design space to a lightweight $<$10K-parameter linear router and suggesting future work should invest in decomposition quality over router expressiveness. On a bandwidth-constrained GTX\,960 (112\,GB/s), CLEAR-MoE's FFN is 1.3--1.7$\times$ slower than dense because loading shared weights for all tokens dominates; roofline analysis confirms routing/scatter-gather (AI\,=\,1.0--1.9) are 11--21$\times$ below the ridge point while expert GEMMs (AI\,=\,22.7) approach the compute ceiling, identifying fused dispatch kernels as the primary engineering target. Future research directions include evaluating scaling on ViT-L and ImageNet-1K, implementing Triton-fused router-dispatch kernels, extending the composite scoring to hierarchical backbones (Swin, ConvNeXt) with non-uniform FFN widths, and empirically testing CLEAR-MoE on high-bandwidth hardware (A100/H100 class) to validate projected latency parity.

%â”€â”€â”€â”€â”€â”€â”€â”€â”€â”€â”€â”€â”€â”€â”€â”€â”€â”€â”€â”€â”€â”€â”€â”€â”€â”€â”€â”€â”€â”€â”€â”€â”€â”€â”€â”€â”€â”€â”€â”€â”€â”€â”€â”€â”€â”€â”€â”€â”€â”€â”€â”€â”€â”€â”€â”€â”€â”€â”€â”€â”€â”€â”€â”€â”€â”€â”€â”€â”€â”€â”€â”€â”€â”€â”€â”€â”€
% \IEEEtriggeratref{N}  % uncomment and tune N if refs still spill to page 7 after removing \clearpage
\balance
\bibliographystyle{IEEEtran}
\bibliography{references}

%â”€â”€â”€â”€â”€â”€â”€â”€â”€â”€â”€â”€â”€â”€â”€â”€â”€â”€â”€â”€â”€â”€â”€â”€â”€â”€â”€â”€â”€â”€â”€â”€â”€â”€â”€â”€â”€â”€â”€â”€â”€â”€â”€â”€â”€â”€â”€â”€â”€â”€â”€â”€â”€â”€â”€â”€â”€â”€â”€â”€â”€â”€â”€â”€â”€â”€â”€â”€â”€â”€â”€â”€â”€â”€â”€â”€â”€
%  APPENDIX
%â”€â”€â”€â”€â”€â”€â”€â”€â”€â”€â”€â”€â”€â”€â”€â”€â”€â”€â”€â”€â”€â”€â”€â”€â”€â”€â”€â”€â”€â”€â”€â”€â”€â”€â”€â”€â”€â”€â”€â”€â”€â”€â”€â”€â”€â”€â”€â”€â”€â”€â”€â”€â”€â”€â”€â”€â”€â”€â”€â”€â”€â”€â”€â”€â”€â”€â”€â”€â”€â”€â”€â”€â”€â”€â”€â”€â”€
\clearpage
\appendices
% Appendix tables/figures labeled A1, A2, B1... (section letter + local counter)
\makeatletter
\@addtoreset{table}{section}
\@addtoreset{figure}{section}
\makeatother
\renewcommand{\thetable}{\thesection\arabic{table}}
\renewcommand{\thefigure}{\thesection\arabic{figure}}

\section{Methodology Details}
\label{app:method_details}

\subsection{Glossary of Key Terms}
\label{app:glossary}

Table~\ref{tab:glossary} defines domain-specific terminology used throughout the paper.

\begin{table}[H]
\caption{Domain-specific terminology.}
\label{tab:glossary}
\centering
\small
\renewcommand{\arraystretch}{1.05}
\resizebox{\columnwidth}{!}{%
\begin{tabular}{p{2.0cm}p{5.4cm}}
\toprule
\textbf{Term} & \textbf{Definition} \\
\midrule
Token & Image patch embedded as $d$-dimensional vector; 197 tokens per $224{\times}224$ image \\
FFN & Feed-forward network: two-layer MLP applied per token in each transformer block \\
Expert & Routed residual branch processing a token subset; in CLEAR-MoE, all residual experts share one direction $(W_2{-}W_\text{shared})$ and differ only by a cluster-conditioned scalar $s_e$ \\
Router/Gate & Lightweight network mapping each token to a probability distribution over experts \\
MoE layer & FFN replaced by $E$ experts plus a router; only top-$k{<}E$ experts activate per token \\
SVD & Singular value decomposition; truncated SVD retains top-$r$ singular components \\
$k$-means & Partitions $N$ points into $k$ clusters by minimising within-cluster variance \\
Calibration set & Small representative set used to measure model statistics (no weight updates) \\
AI & Arithmetic intensity: FLOPs per byte of memory accessed \\
Ridge point & AI where compute ceiling equals memory bandwidth ceiling (21.4 FLOPs/B on GTX\,960) \\
p50 Latency & Median over 100 repeated forward passes; robust to OS-scheduling outliers \\
cuBLAS & NVIDIA's optimised linear-algebra library; batched GEMM calls dispatch through it \\
\bottomrule
\end{tabular}
}
\end{table}

\FloatBarrier
\subsection{Dataset Preparation and EDA}
\label{app:eda}

\textbf{Imagenette.} 13,394 raw images were scanned using SHA-256 deduplication, robust z-score filtering ($\tau{=}3.5$), and Isolation Forest (contamination\,=\,0.01). 118 images (0.88\%) were removed, yielding a clean set of 9,391 train / 3,885 val (used for EDA analysis only). The \emph{standard} Imagenette validation split (3,925 images, including the 40 removed as anomalies by our filter) was used for all reported Top-1 accuracy evaluation, ensuring comparability with prior work. ImageNet normalisation and Lanczos-4 resize to $224{\times}224$ are applied.

\textbf{Distribution shift analysis.} DeiT-Small penultimate activations for all 3,885 clean val images (shape $3885{\times}384$): PSI\,=\,0.043 ($<$0.1, negligible shift), JSD\,=\,0.018, KS $p$-values\,$>$0.05 for all 10 classes. \textbf{Clustering quality.} $k{=}10$ on penultimate image-level activations: Silhouette\,=\,0.542, Davies-Bouldin\,=\,1.24, confirming that the feature space supports class-level cluster separation. Note that CLEAR-MoE clusters token-level intermediate activations with $k{=}4$, which is a complementary but distinct analysis; the image-level EDA establishes feature quality rather than directly validating token cluster assignments. \textbf{SVD energy.} Rank-192 truncation retains 99.2\% of penultimate activation matrix variance; top-10 singular values capture 67.8\%. The effect of $W_2$ truncation rank on accuracy and reconstruction error is evaluated in Table~\ref{tab:rank_sweep}.

Table~\ref{tab:eda} summarises these preprocessing and analysis statistics.

\begin{table}[H]
\caption{Dataset preprocessing and EDA statistics.}
\label{tab:eda}
\centering
\small
\resizebox{\columnwidth}{!}{%
\begin{tabular}{p{3.5cm}p{4.0cm}}
\toprule
\textbf{Statistic} & \textbf{Value} \\
\midrule
Raw Imagenette images & 13,394 \\
Removed (anomalous) & 118 (0.88\%) \\
Train / val split & 9,391 / 3,885 (clean, EDA only); 3,925 (standard, used for all Top-1) \\
Train-test PSI & 0.043 ($<$0.1 = negligible shift) \\
Train-test JSD & 0.018 \\
$k{=}10$ Silhouette score & 0.542 \\
$k{=}10$ Davies-Bouldin & 1.24 \\
Rank-192 SVD variance & 99.2\% \\
Calibration size & 200 images \\
\bottomrule
\end{tabular}
}
\end{table}

\FloatBarrier
\subsection{Algorithm Pseudocode}
\label{app:algorithms}

Algorithms~\ref{alg:scoring}--\ref{alg:inference} give pseudocode for the three core CLEAR-MoE phases.

\begin{algorithm}[H]
\caption{Phase 2: Layer Scoring and Selection}
\label{alg:scoring}
\begin{algorithmic}[1]
\REQUIRE Calibration activations $\{h_l^{(i)}\}$, $L$\,=\,12 layers, select $k{=}6$
\FOR{each layer $l$}
  \STATE $\text{sp}(l) \gets \frac{\sum_j \mathbf{1}(|h_{l,j}|<0.01)}{\operatorname{numel}(h_l)}$
  \STATE $\text{cl}(l) \gets \text{Silhouette}(k\text{-means}(h_l, E))$
  \STATE $\text{se}(l) \gets \|\Delta\text{logits}\|/\|\text{logits}\|$ \hfill (zero FFN output)
  \STATE $\mathcal{S}(l) \gets 0.4\,\text{sp} + 0.4\,\text{cl} - 0.2\,\text{se}$
\ENDFOR
\STATE Select top-$k$ layers by $\mathcal{S}$
\end{algorithmic}
\end{algorithm}

\begin{algorithm}[H]
\caption{Phase 3: Expert Extraction (fc1 shared; fc2 decomposed)}
\label{alg:extraction}
\begin{algorithmic}[1]
\REQUIRE $W_2 \in \mathbb{R}^{d \times d_\text{ffn}}$, activations $z$, $E$, rank $r$
\STATE $[U,\Sigma,V] \gets \text{SVD}(W_2)$; \; $W_\text{shared} \gets U_r\Sigma_r V_r^\top$
\STATE $\{C_1,\ldots,C_E\} \gets k\text{-means}(z, E)$
\FOR{$e \in \{1,\ldots,E\}$}
  \STATE $s_e \gets \frac{\frac{1}{|C_e|}\sum_{i\in C_e}\|z_i\|_2}{\frac{1}{N}\sum_{i=1}^{N}\|z_i\|_2}$
  \STATE $W_e^\text{res} \gets (W_2 - W_\text{shared}) \cdot s_e$
\ENDFOR
\end{algorithmic}
\end{algorithm}

\begin{algorithm}[H]
\caption{CLEAR-MoE FFN Inference (Grouped backend)}
\label{alg:inference}
\begin{algorithmic}[1]
\REQUIRE Tokens $x$, shared fc1, $W_\text{shared}$, $\{W_e^\text{res}\}$, Router $g$
\STATE $z \gets \text{GELU}(\text{fc1}(x))$ \hfill (shared, all tokens)
\STATE $h_\text{sh} \gets z W_\text{shared}^\top$ \hfill (shared SVD basis)
\STATE $a \gets \text{argmax}(g(x))$ \hfill (expert assignment)
\STATE Sort $z$ by $a$; get sub-batch boundaries via index search
\STATE $h_\text{res}[e] \gets z_e W_e^{\text{res}\top}$ for each expert $e$
\STATE Unshuffle $h_\text{res}$; $y \gets h_\text{sh} + h_\text{res}$
\end{algorithmic}
\end{algorithm}

\FloatBarrier
\section{Experimental Configuration}
\label{app:experimental_config}

\subsection{Hardware and Software Setup}
\label{app:setup}

Table~\ref{tab:setup} lists the complete hardware and software configuration used in all experiments.

\begin{table}[H]
\caption{Hardware and software configuration for all experiments.}
\label{tab:setup}
\centering
\small
\resizebox{\columnwidth}{!}{%
\begin{tabular}{p{3.0cm}p{4.5cm}}
\toprule
\textbf{Item} & \textbf{Specification} \\
\midrule
OS & Windows 11 Pro \\
GPU & NVIDIA GeForce GTX\,960 \\
GPU memory & 4.0\,GB GDDR5 \\
Memory bandwidth & 112\,GB/s \\
Peak FP32 throughput & 2.4\,TFLOP/s \\
Ridge point & 21.4\,FLOPs/Byte \\
CUDA & 11.8 \\
PyTorch & 2.6.0+cu118 \\
Classification backbone & DeiT-Small; 22M parameters; patch size 16$\times$16; $d{=}384$; $d_\text{ffn}{=}1536$; 12 transformer blocks \\
Segmentation backbone & Not evaluated (planned future work) \\
Default $E$ & 4 experts \\
Default rank $r$ & $r{=}d/2{=}192$ for DeiT-Small (50\% of matrix rank) \\
Calibration size & 200 images \\
Router training & 5 epochs, AdamW, lr\,$10^{-3}$, cosine decay \\
Latency & p50 of 100 passes, batch\,=\,8, \texttt{cuda.synchronize()} \\
\bottomrule
\end{tabular}
}
\end{table}

\FloatBarrier
\section{Extended Experimental Results}
\label{app:extended_results}

\subsection{Full Ablation Tables}
\label{app:ablation}

\textbf{Per-layer composite scores.} Full composite scores for all 12 DeiT-Small FFN blocks appear in Table~\ref{tab:layer_scores_main} (Section~\ref{sec:method}). Block~0 ranks last (composite\,=\,0.101) due to sensitivity\,=\,0.946.

\textbf{Full layer-selection policy ablation.} The complete L0--L10 ablation now appears in Table~\ref{tab:ablation} (Section~\ref{sec:ablation}).

\FloatBarrier
\subsection{Hyperparameter Sensitivity Studies}
\label{app:sensitivity}

Tables~\ref{tab:seeds}--\ref{tab:router_comparison} report the full numerical results for the five hyperparameter sensitivity studies summarised in Section~\ref{sec:results}.

\begin{table}[H]
\caption{3-Seed Robustness. Full composite-score pipeline, $E{=}4$, DeiT-Small.}
\label{tab:seeds}
\centering
\small
\resizebox{\columnwidth}{!}{%
\begin{tabular}{lrrrr}
\toprule
\textbf{Seed} & \textbf{Top-1} & \textbf{p50 ms} & \textbf{Router Acc} & \textbf{Skew} \\
\midrule
42  & 86.68\% & 85.05 & 0.925 & 0.099 \\
123 & 86.70\% & 83.90 & 0.928 & 0.116 \\
456 & 86.73\% & 78.89 & 0.928 & 0.101 \\
\midrule
\textbf{Mean$\pm$Std} & \textbf{86.70$\pm$0.02\%} & 82.6$\pm$2.7 & 0.927$\pm$0.002 & 0.105$\pm$0.008 \\
\bottomrule
\end{tabular}
}
\end{table}

\begin{table}[H]
\caption{Calibration-set sensitivity. Multiple random subsets per size.}
\label{tab:calib}
\centering
\small
\resizebox{\columnwidth}{!}{%
\begin{tabular}{rrrrrr}
\toprule
$N_\text{cal}$ & \textbf{Sub.} & \textbf{Top-1} & \textbf{Router Acc} & \textbf{Skew} \\
\midrule
 50 & 1 & 86.70\% & 0.867 & 0.139 \\
 50 & 2 & 86.68\% & 0.833 & 0.073 \\
 50 & 3 & 86.65\% & 0.856 & 0.139 \\
100 & 1 & 86.55\% & 0.896 & 0.094 \\
100 & 2 & 86.60\% & 0.908 & 0.150 \\
100 & 3 & 86.70\% & 0.899 & 0.153 \\
200 & 1 & 86.68\% & 0.926 & 0.099 \\
200 & 2 & 86.60\% & 0.928 & 0.128 \\
500 & 1 & 86.70\% & 0.953 & 0.098 \\
\bottomrule
\end{tabular}
}
\end{table}

\begin{table}[H]
\caption{SVD rank sweep. Seed~42, $E{=}4$, composite scoring.}
\label{tab:rank_sweep}
\centering
\small
\begin{tabular}{rrrr}
\toprule
\textbf{Rank $r$} & \textbf{Top-1} & \textbf{p50 ms} & \textbf{Recon Err} \\
\midrule
 16 & 86.75\% & 78.13 & 0.907 \\
 32 & 86.65\% & 76.79 & 0.841 \\
 64 & 86.65\% & 76.60 & 0.734 \\
 96 & 86.75\% & 78.07 & 0.643 \\
128 & 86.68\% & 77.82 & 0.562 \\
192 & 86.62\% & 76.01 & 0.420 \\
256 & 86.70\% & 77.36 & 0.294 \\
\bottomrule
\end{tabular}
\end{table}

\begin{figure}[H]
\centering
\includegraphics[width=0.85\columnwidth]{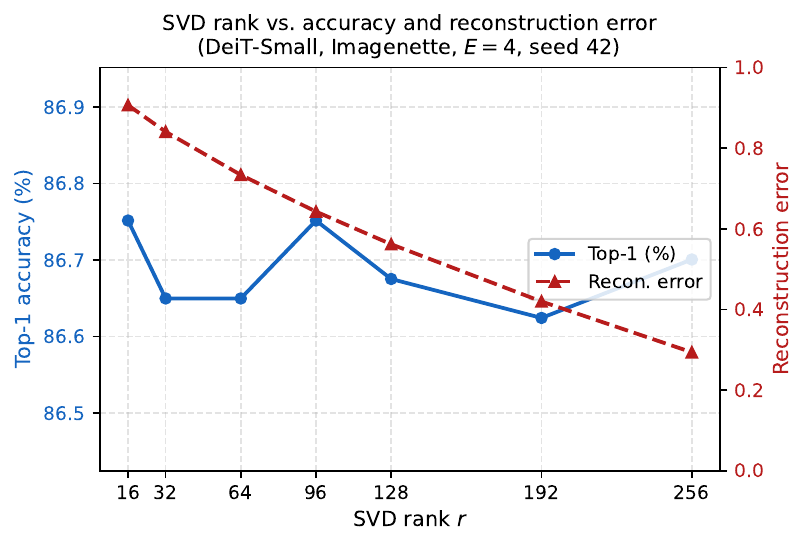}
\caption{SVD rank vs.\ Top-1 accuracy (left axis, blue) and reconstruction error (right axis, red). Accuracy spans 0.13\,pp across $r\in\{16,\ldots,256\}$ while reconstruction error drops from 0.907 to 0.294, confirming rank does not gate accuracy once the shared basis captures the dominant singular subspace.}
\label{fig:rank_sweep}
\end{figure}

\begin{table}[H]
\caption{Expert-count study. Seed~42, composite scoring, $k{=}6$.}
\label{tab:expert_count}
\centering
\small
\begin{tabular}{rrrrrr}
\toprule
$E$ & \textbf{Top-1} & \textbf{p50 ms} & \textbf{Rtr Acc} & \textbf{Skew} & \textbf{Empty} \\
\midrule
 2 & 86.65\% & 73.4 & 0.967 & 0.247 & 0 \\
 4 & 86.65\% & 77.5 & 0.923 & 0.074 & 0 \\
 8 & 86.80\% & 84.3 & 0.893 & 0.042 & 0 \\
16 & 86.39\% & 97.8 & 0.865 & 0.024 & 0 \\
\bottomrule
\end{tabular}
\end{table}

\begin{table}[H]
\caption{Router architecture comparison. Seed~42, $E{=}4$, composite scoring. Entropy: aggregate expert-usage entropy (distribution of tokens across experts, per layer; max $=\ln 4 \approx 1.386$). Note: this is the \emph{load-balance} entropy, distinct from the per-token predictive entropy in Fig.~\ref{fig:expert_entropy} ($\approx$0.40 nats), which measures router confidence.}
\label{tab:router_comparison}
\centering
\small
\resizebox{\columnwidth}{!}{%
\begin{tabular}{lrrrrrrr}
\toprule
\textbf{Router} & \textbf{Top-1} & \textbf{p50 ms} & \textbf{Rtr Acc} & \textbf{Skew} & \textbf{Entropy} & \textbf{Params} \\
\midrule
Linear   & 86.68\% & 78.0 & 0.925 & 0.099 & 1.296 & 9,240 \\
MLP      & 86.68\% & 75.6 & 0.980 & 0.095 & 1.300 & 149,400 \\
Adaptive & 86.68\% & 77.6 & 0.936 & 0.099 & 1.296 & 9,240 \\
\bottomrule
\end{tabular}
}
\end{table}

\FloatBarrier
\subsection{Dispatch Benchmark and Parallel Scaling}
\label{app:dispatch}

Table~\ref{tab:dispatch} benchmarks all three CUDA backends under four token-load imbalance levels ($B{=}8$, $T{=}1568$ tokens, $E{=}4$, GTX\,960); CPU Serial is included as a baseline, allowing direct GPU-vs-CPU comparison at each imbalance level. The roofline plot appears in Fig.~\ref{fig:roofline} (Section~\ref{sec:results}). Table~\ref{tab:parallel} projects multi-device throughput analytically.

\begin{table}[H]
\caption{Dispatch micro-benchmark: p50 latency (ms) and throughput (K\,tok/s) across imbalance levels. $T{=}1568 = 8{\times}196$ tokens (class token excluded from expert dispatch; processed on the dense path). Calibration uses 197 tokens/image including the class token.}
\label{tab:dispatch}
\centering
\footnotesize
\setlength{\tabcolsep}{3pt}
\resizebox{\columnwidth}{!}{%
\begin{tabular}{lrrrrrrrr}
\toprule
& \multicolumn{2}{c}{\textbf{0\% imb.}} & \multicolumn{2}{c}{\textbf{40\% imb.}} & \multicolumn{2}{c}{\textbf{60\% imb.}} & \multicolumn{2}{c}{\textbf{80\% imb.}} \\
\cmidrule(lr){2-3}\cmidrule(lr){4-5}\cmidrule(lr){6-7}\cmidrule(lr){8-9}
\textbf{Backend} & ms & K/s & ms & K/s & ms & K/s & ms & K/s \\
\midrule
CPU Serial & 6.01 & 261 & 7.22 & 217 & 5.96 & 263 & 5.96 & 263 \\
Naive       & 3.67 & 428 & 3.55 & 442 & 3.50 & 448 & 3.59 & 437 \\
Grouped     & 2.36 & 665 & 2.10 & 747 & 2.30 & 681 & 2.14 & 733 \\
cuBLAS      & \textbf{2.15} & \textbf{728} & 2.29 & 684 & 2.45 & 639 & 2.81 & 558 \\
\bottomrule
\end{tabular}
}
\end{table}

\begin{figure}[H]
\centering
\includegraphics[width=0.85\columnwidth]{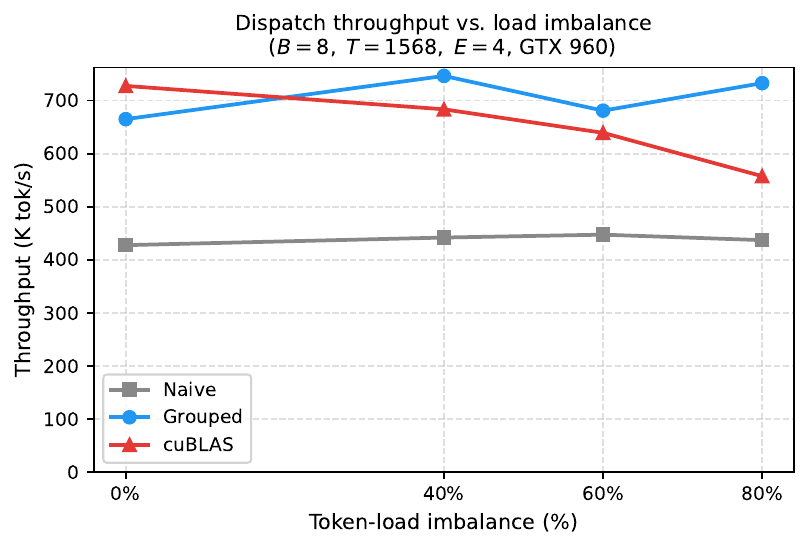}
\caption{Dispatch throughput vs.\ token-load imbalance for three CUDA backends ($B{=}8$, $T{=}1568$, $E{=}4$, GTX\,960). Grouped is stable across all imbalance levels (665--747\,K\,tok/s); cuBLAS degrades 23\% from 0\% to 80\% imbalance due to padding waste. For variable-load workloads, Grouped is the recommended backend.}
\label{fig:dispatch_curve}
\end{figure}

\textbf{GPU vs.\ CPU and Parallel Scaling.}
\label{app:parallel}

GPU vs.\ CPU speedup at balanced load is readable from the CPU Serial row in Table~\ref{tab:dispatch} above: at 0\% imbalance, cuBLAS achieves 728\,K\,tok/s vs.\ 261\,K\,tok/s for CPU Serial ($2.79\times$ speedup).

\textbf{Multi-device projection (simulated; no second GPU available).} All-to-all and AllReduce costs are modelled analytically using PCIe Gen3$\times$16 bandwidth (16\,GB/s). Results are \emph{not} measured NCCL runs.

\begin{table}[H]
\caption{Parallel scaling projection. Expert/Data/Pipeline parallelism are simulated; single-GPU is measured.}
\label{tab:parallel}
\centering
\small
\resizebox{\columnwidth}{!}{%
\begin{tabular}{lrrrr}
\toprule
\textbf{Mode} & \textbf{tok/s} & \textbf{Devices} & \textbf{Speedup vs GPU-1} & \textbf{Efficiency} \\
\midrule
Full-model single-GPU  & 225K & 1 & 1.00$\times$ & 1.00 \\
Expert Parallel EP-2 (sim.) & 175K & 2 & 0.78$\times$ & 0.39 \\
Pipeline Parallel PP-2 (sim.) & 8K & 2 & 0.03$\times$ & 0.02 \\
\bottomrule
\multicolumn{5}{l}{\footnotesize All multi-device entries are simulated via PCIe Gen3$\times$16 bandwidth; no NCCL runs performed.} \\
\multicolumn{5}{l}{\footnotesize EP-2 overhead: all-to-all token redistribution (AI\,=\,0.1\,F/B).} \\
\multicolumn{5}{l}{\footnotesize PP-2: 50\% pipeline bubble with 1 micro-batch, 2 stages.} \\
\multicolumn{5}{l}{\footnotesize Single-GPU throughput is full-model (distinct from dispatch-only throughput in Table~C6).} \\
\end{tabular}
}
\end{table}

\end{document}